\documentclass[conference,letterpaper]{IEEEtran}
\IEEEoverridecommandlockouts
\usepackage{cite}
\usepackage{amsmath,amssymb,amsfonts}
\usepackage{algorithmic}
\usepackage{algorithm}
\usepackage{graphicx}
\usepackage{textcomp}
\usepackage{xcolor}
\usepackage{array}
\usepackage[caption=false,font=normalsize,labelfont=sf,textfont=sf]{subfig}
\usepackage{url}
\usepackage{booktabs}
\usepackage{multirow}

\def\BibTeX{{\rm B\kern-.05em{\sc i\kern-.025em b}\kern-.08em
    T\kern-.1667em\lower.7ex\hbox{E}\kern-.125emX}}

\begin{document}

\title{FilterFusion: Prior-Guided Adaptive Filter Fusion for Efficient Video Denoising}

\author{
\IEEEauthorblockN{%
Wenlong Yang\IEEEauthorrefmark{1}\IEEEauthorrefmark{2},\
Weihang Yuan\IEEEauthorrefmark{2},\
Canran Jin\IEEEauthorrefmark{2},\
Jiahao Li\IEEEauthorrefmark{2},\
Henrik Turbell\IEEEauthorrefmark{2},\
Lifeng Sun\IEEEauthorrefmark{1}}
\IEEEauthorblockA{\IEEEauthorrefmark{1}Tsinghua University, Beijing, China\\
\IEEEauthorrefmark{2}Microsoft\\
\texttt{ywl22@mails.tsinghua.edu.cn}, \texttt{sunlf@tsinghua.edu.cn}}
}

\maketitle

\begin{abstract}
Real-world video denoising for communication systems must handle mixed degradations such as sensor noise, compression artifacts, and motion-induced inconsistency while remaining deployable under tight latency budgets. Purely deep models often achieve high accuracy but come with significant computational cost, whereas classical filters are robust and efficient but insufficiently adaptive. We present \textbf{FilterFusion}, a prior-guided multi-frame denoising framework that combines temporally-aware learned features with an interpretable bank of denoising branches. The model predicts a low-cost guidance signal and uses it to adaptively mix a compact set of denoising responses at the original resolution. The lightweight design uses only 109K parameters and explicitly injects temporal variance and spatial variance cues into the low-resolution controller. On CRVD (real sensor noise), it reaches 34.72 dB PSNR and 0.9330 SSIM with only 0.18 GFLOPs (network-only). Because the filter-bank operations are simple and regular (Gaussian blur, median, weighted sum), a straightforward compiled C implementation of the synthesis path achieves 17.9\,ms per frame (55.9\,FPS) on a single CPU thread at 1080p---exceeding real-time and $122{\times}$--$252{\times}$ faster than lightweight CNN baselines that require over 2 seconds per frame. The results demonstrate a strong quality-efficiency-speed trade-off and a practical route to real-time denoising through low-resolution control and explicit filter-based synthesis.
\end{abstract}

\begin{IEEEkeywords}
video denoising, real-noise denoising, adaptive filter fusion, prior-guided restoration, lightweight video enhancement
\end{IEEEkeywords}

\section{Introduction}

Video denoising is a core component in video communication, live streaming, and low-light capture. In these scenarios, degradations are rarely limited to additive white Gaussian noise; instead, sensor noise, gain amplification, compression artifacts, and motion interact in a spatially and temporally non-uniform manner. This makes real-world denoising substantially harder than synthetic benchmark restoration.

\begin{figure}[!t]
\centering
\includegraphics[width=0.48\textwidth]{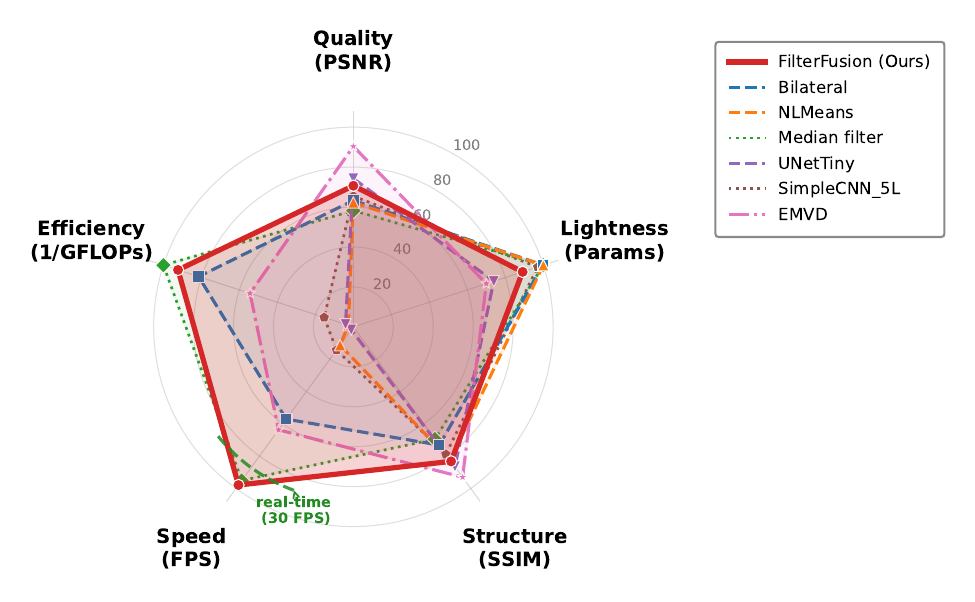}
\caption{Multi-dimensional trade-off. Each axis is normalized to $[0,1]$: \textbf{Quality}~(PSNR, $[22,40]$\,dB), \textbf{Lightness}~($1-\text{params}/10^{6}$; classical${=}$max), \textbf{Structure}~(SSIM, $[0.6,1.0]$), \textbf{Speed}~($\log$ FPS), \textbf{Efficiency}~($\log\,1/\text{GFLOPs}$). The dashed green arc marks the real-time threshold (30\,FPS). FilterFusion (red) is the only method that simultaneously exceeds the real-time boundary and surpasses the classical quality ceiling.}
\label{fig:radar}
\end{figure}

Classical video denoisers exploit temporal redundancy and robust filtering priors, and therefore often behave predictably in deployment~\cite{dabov2007video,arias2018video,maggioni2012video}. However, they rely on hand-crafted rules and struggle to adapt to diverse local content. Deep learning methods, in contrast, can learn rich restoration mappings~\cite{tassano2019dvdnet,tassano2020fastdvdnet,chan2022basicvsrpp,yue2020supervised}, but their performance may drop sharply when the test degradation deviates from the training distribution. In addition, fully black-box designs are difficult to analyze and to compress for practical systems.

This paper revisits video denoising from a structured design perspective. Instead of forcing a single backbone to learn all restoration behavior implicitly, we explicitly construct several branches with distinct denoising roles and let the network decide how much each branch should contribute at every pixel. The proposed \textbf{FilterFusion} framework combines a temporal feature backbone with a prior-guided filter bank, and further compresses the control path into a low-resolution weight prediction module. A compact prediction head generates pixel-wise fusion weights, enabling content-aware and noise-aware selection among complementary operators while avoiding expensive full-resolution neural decoding.

The key advantage of this design is that it retains the efficiency, interpretability and cross-degradation robustness of classical filtering while benefiting from learned content-aware adaptivity. The controller assigns responses per pixel: multi-frame temporal averaging that directly suppresses noise in motion-stable regions, multi-scale low-pass smoothing for residual structure on top of averaged signal, median rejection near compression artifacts, and reference-frame preservation across edges, textures, and motion-unreliable areas.

\textbf{Contributions.} Our main contributions are as follows:
\begin{itemize}
    \item We propose FilterFusion, a structured video denoising framework that embeds classical filtering priors into a learnable multi-branch architecture rather than treating denoising as a monolithic black-box mapping.
    \item We instantiate the framework as a 109K-parameter multi-frame model and introduce explicit temporal variance and spatial variance guidance for robust and compact low-resolution weight prediction.
    \item We provide detailed real-noise and CPU-side comparisons on CRVD (sensor noise) where the model reaches 34.72 dB with only 0.51 GFLOPs total (0.18 network-only) and shows a favorable quality-efficiency trade-off, requiring $120{\times}$--$240{\times}$ less compute than matched-quality full-resolution CNN baselines.
\end{itemize}

\section{Related Work}

\textbf{Video Denoising.}
Classical methods such as VBM3D and related non-local formulations leverage patch redundancy and temporal aggregation to obtain robust denoising performance~\cite{dabov2007video,arias2018video,maggioni2012video}. More recent learning-based methods, including DVDnet, FastDVDnet, RViDeNet, and recurrent propagation models, have substantially improved restoration quality on synthetic and curated datasets~\cite{tassano2019dvdnet,tassano2020fastdvdnet,yue2020supervised,chan2022basicvsrpp}. Their main weakness is high computational cost and limited interpretability of the denoising decision in mixed-noise scenarios.

\textbf{Classical Priors in Learnable Restoration.}
Many efficient restoration models borrow ideas from classical filtering, residual learning, and multi-scale processing~\cite{zhang2017beyond,zhang2018ffdnet,liu2020residual,zamir2022restormer}. Recent hybrid approaches explicitly combine classical denoising modules with learned predictors~\cite{jin2025classicvideodenoisingmachine}. Our work follows this line of thought but emphasizes a particularly deployment-friendly form: most adaptive decisions are made in a low-resolution latent space, while full-resolution processing is reduced to a small bank of interpretable filter responses.

\section{Approach}
\label{sec:approach}

\begin{figure}[!t]
    \centering
    \includegraphics[width=0.48\textwidth]{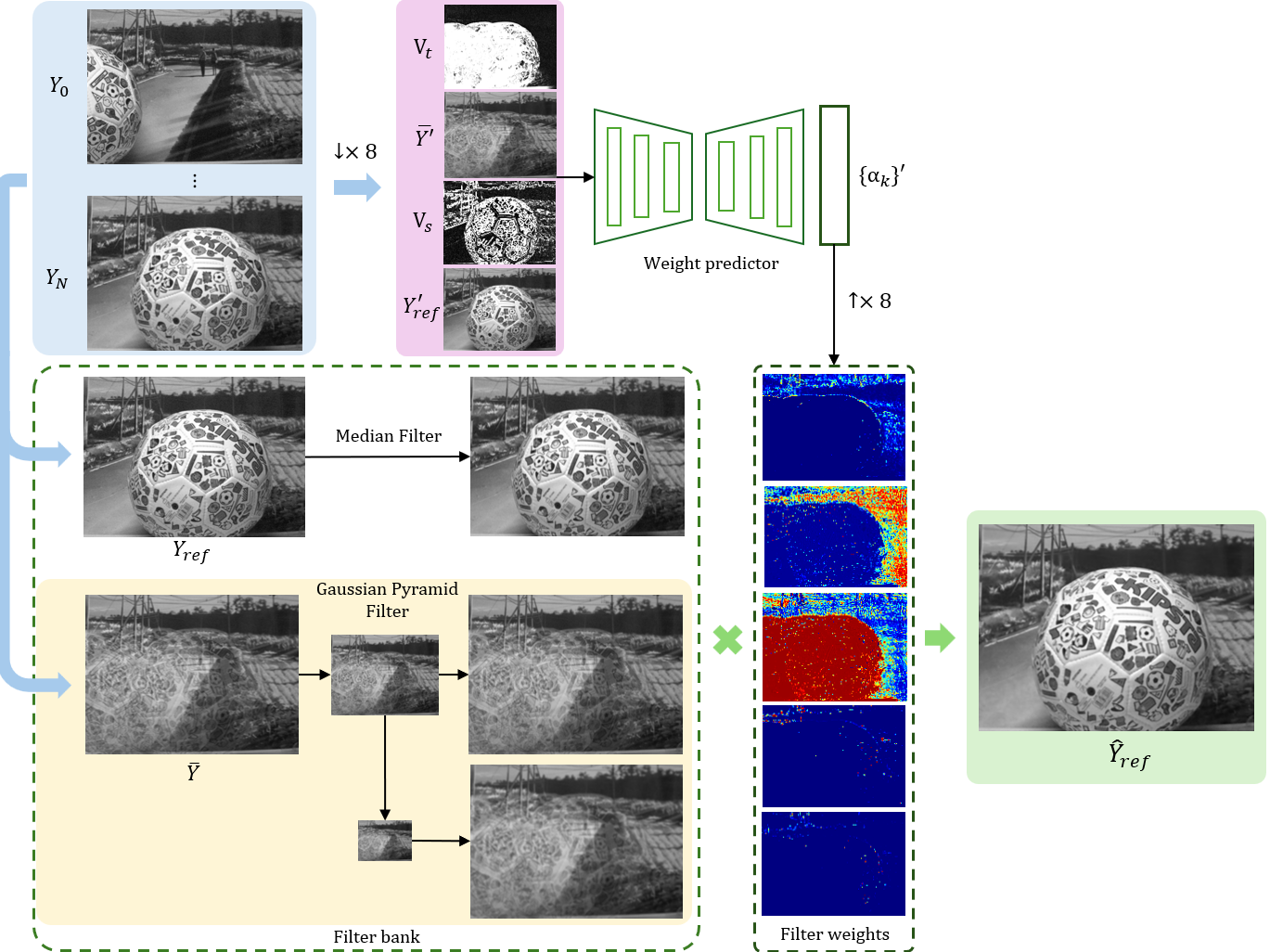}
    \caption{Overview of FilterFusion in luma plane. A compact low-resolution predictor estimates pixel-wise fusion weights from the reference frame \(Y_{\mathrm{ref}}\), temporal mean \(\bar{Y}\), and variance guidance \((V_s, V_t)\), while a full-resolution filter bank generates multiple candidate responses. The final prediction is obtained by adaptively fusing these responses using the predicted weights. For the chroma planes, the same filter-bank synthesis is applied while reusing the weights predicted from the luma plane, producing the denoised NV12 output with negligible additional cost.}
    \label{fig:overview}
\end{figure}

Given a temporal window of $N$ frames $\{ I_n \}_{n=1}^{N}$, our goal is to reconstruct a clean reference frame $I_{\mathrm{ref}}$ where usually $I_{\mathrm{ref}}=I_N$. Figure~\ref{fig:overview} illustrates the pipeline. FilterFusion first extracts several compact low-resolution guidance maps from the input sequence and then uses these guidance to fuse several full-resolution denoising responses. This decomposition separates low-cost adaptive decision making from full-resolution image synthesis and is the core reason the method remains computationally practical.

\subsection{Problem Formulation}

 
Let $\hat{Y}_{\mathrm{ref}}$ denote the denoised luma channel of the reference frame; chroma is handled by reusing the same weights, as detailed in the synthesis subsection. We formulate restoration as adaptive mixing over a filter bank:
\begin{equation}
\label{eq:fusion}
\hat{Y}_{\mathrm{ref}}(x,y)=\sum_{k=1}^{K} \alpha_k(x,y) B_k(x,y),
\end{equation}
where $B_k$ is the $k$-th filter-bank response on the luma plane and $\{\alpha_k\}$ is its pixel-wise mixing weight. The key question is how to estimate $\alpha_k$ with minimal cost while keeping the basis functions expressive enough to cover different local noise conditions. Our answer is to predict $\{\alpha_k\}$ in low-resolution as $\{\alpha_k\}'$ and keep the basis functions at the original resolution.

\subsection{Low-Resolution Weight Prediction}

The model uses a compact low-resolution predictor to estimate the spatially varying fusion weights $\{\alpha_k\}'$. Structurally, this predictor is a lightweight encoder-decoder that takes downsampled noisy frames and outputs a small set of fusion logits. The model contains about 109K parameters, so the learned part remains compact compared with conventional full-resolution video restoration networks. In the current implementation, instead of feeding all $N$ frames into the predictor, we only use two frames to supply both spatial and temporal information at minimal computation: reference frame $Y_{\mathrm{ref}}$ and the temporal mean $\bar{Y}=\tfrac{1}{N}\sum_{n=1}^{N} Y_n$. Both are taken from the Y channel which carries the dominant structural information of the frame, and bilinearly downsampled by $8\times$ before entering the predictor. This moves the learned computation away from both the full-resolution domain and the full temporal window.

The predictor outputs a 5-channel logit map in the low-resolution domain, where each channel corresponds to one of the five synthesis responses of the filter bank. After softmax normalization, the predicted guidance is upsampled back to the original frame size and used for full-resolution synthesis. This architecture is motivated by a simple observation: the spatial variation of optimal filter-selection weights is much smoother than the final image itself, so the control signal can be inferred at low resolution without sacrificing the main adaptive behavior. To make low-resolution prediction reliable under motion and structural variation, the predictor is conditioned on two compact guidance signals introduced below.

\textbf{Temporal Variance Guidance.} The temporal condition addresses frame-to-frame inconsistency caused by motion, illumination changes, or flickering. In such regions, blind multi-frame fusion risks ghosting artifacts because the controller has no explicit signal about temporal reliability. To resolve this, we compute a per-pixel temporal variance map across the input window and then apply nearest-neighbor downsampling:
\begin{equation}
V_t = \mathcal{D}_{\mathrm{NN}}\!\left(\mathrm{Var}\bigl[Y_n(x,y)\bigr]\right),
\end{equation}
where $Y_n$ is the luminance channel of the $n$-th frame in the temporal window. Although the denoising network takes only a compact two-frame input, $V_t$ is computed from all frames in the window and provides a lightweight temporal-consistency cue that this compact input alone cannot convey, without requiring an explicit motion estimation module. Pixels with larger $V_t
$ are guided to rely more on reference-preserving responses rather than aggressive temporal fusion.

\textbf{Spatial Variance Guidance.} The spatial condition addresses a separate problem: after 8$\times$ downsampling, each low-resolution pixel corresponds to an $8{\times}8$ high-resolution region that may be flat or rich in texture. Bilinear downsampling retains only the mean and discards internal complexity, so the controller cannot distinguish flat-noise areas (where strong smoothing helps) from edge-texture areas (where over-smoothing should be avoided).

To supply this information at negligible cost, spatial variance is computed directly in the low-resolution domain using $M{=}9$ positions $\{p_k\}_{k=1}^{9}$ uniformly sampled from the $8{\times}8$ high-resolution coverage region of each low-resolution pixel $(i,j)$ via a $3{\times}3$ sub-pixel grid:
\begin{equation}
V_s(i,j)=\frac{1}{K}\sum_{m=1}^{M}Y_{\mathrm{ref}}^{(m)\,2}(i,j)-\left(\frac{1}{K}\sum_{m=1}^{M}Y_{\mathrm{ref}}^{(m)}(i,j)\right)^2.
\end{equation}
Low $V_s$ indicates flat regions (sky, walls) where stronger smoothing is safe; high $V_s$ indicates structures that should be preserved.

The model input is constructed as follows. The reference frame $Y_\mathrm{ref}$ and temporal mean $\bar{Y}$ are also $8{\times}$ downsampled to get $Y_\mathrm{ref}'$ and $\bar{Y}'$, and then concatenated with $V_\mathrm{s}$ and $V_\mathrm{t}$. The two variance conditions operate on separate dimensions—temporal consistency across frames and spatial structure within each frame—and jointly improve the reliability of the predicted fusion map under mixed degradations.

\subsection{High-Resolution Filter-Bank Synthesis}

The final output is synthesized from a small bank of full-resolution responses $\{B_k\}_{k=1}^{K}$: (i) the temporal mean $\bar{Y}$ for multi-frame noise averaging; (ii) the reference frame for high-frequency detail preservation; (iii) two upsampled Gaussian-pyramid levels of $\bar{Y}$ for progressively stronger low-pass smoothing; and (iv) a spatial median on the reference frame for edge-preserving denoising. These five regimes are deliberately heterogeneous, letting the controller route each pixel to whichever response best matches the local degradation rather than commit to a single operator.

Built on $\bar{Y}$ rather than the reference frame, the Gaussian pyramid couples efficient $3{\times}3$ multi-scale blurring with implicit temporal denoising: multi-frame averaging delivers $\sqrt{N}$ noise reduction in motion-stable regions for free, so spatial smoothing acts on a pre-cleaned signal and stays mild enough to preserve textures that a raw-frame equivalent would erase. The median branch covers the complementary regime—motion and high-frequency regions on the reference frame—where temporal averaging ghosts and Gaussian smoothing erase the fine structure.

The fusion module predicts $K$ logits $w_k(x,y)$ and normalizes them with a softmax:
\begin{equation}
\alpha_k(x,y)=\frac{\exp(w_k(x,y))}{\sum_{i=1}^{K}\exp(w_i(x,y))}.
\end{equation}


The weights satisfy $\sum_k \alpha_k(x,y)=1$, so Eq.~\ref{eq:fusion} reduces to a per-pixel convex combination of the $K$ candidate responses $B_1,\dots,B_{K}$.


For the chroma planes, we directly reuse the luma-predicted weights $\{\alpha_k\}$: they are upsampled to the chroma resolution ($1/4$ of luma in NV12) and used to combine a UV-side filter bank constructed in the same form as the luma   bank. This weight sharing exploits the structural alignment between luma and chroma, so chroma synthesis adds negligible cost on top of luma synthesis while preserving color consistency.

\subsection{Training Objective}

We train FilterFusion end-to-end using a weighted combination of pixel reconstruction, robust Charbonnier, Sobel-based regularization, and frequency-domain supervision:
\begin{equation}
\mathcal{L}=\lambda_{1}\mathcal{L}_{\mathrm{L1}}+\lambda_{2}\mathcal{L}_{\mathrm{charb}}+\lambda_{3}\mathcal{L}_{\mathrm{sobel}}+\lambda_{4}\mathcal{L}_{\mathrm{freq}}.
\end{equation}
The Sobel term is applied on the prediction to suppress residual high-frequency noise in smooth regions, while the frequency term aligns global spectral characteristics with the target and helps preserve valid high-frequency components. In our stable training configuration, we use $\lambda_{1}=1.0$, $\lambda_{2}=1.0$, $\lambda_{3}=0.1$, and $\lambda_{4}=0.05$.

Concretely, we use
\begin{equation}
\mathcal{L}_{\mathrm{sobel}}=\|\nabla_x \hat{y}\|_1 + \|\nabla_y \hat{y}\|_1,
\end{equation}
and
\begin{equation}
\mathcal{L}_{\mathrm{freq}}=\|\mathcal{F}(\hat{y})-\mathcal{F}(y)\|_1,
\end{equation}
where $\nabla_x,\nabla_y$ are Sobel operators and $\mathcal{F}(\cdot)$ denotes 2D FFT.

Training uses the CRVD sRGB clean as ground truth. We optimize the model with Adam using $\beta_1=0.9$, $\beta_2=0.99$, an initial learning rate of $10^{-4}$, and a multistep schedule with milestones at 140k, 280k, and 360k iterations. The total training length is 600k iterations with batch size 4 per GPU. During optimization, a 2-pixel border crop is excluded from the loss to reduce boundary artifacts.

\section{Experiments}
\label{sec:experiments}

\subsection{Experimental Setup}

We validate the method on \textbf{CRVD}~\cite{yue2020supervised}, a real sensor-noise dataset containing 11 indoor scenes with substantial object motion across 5 ISO levels (1600--25600) at $1920{\times}1080$. Following the official two-stage protocol, we pretrain on synthetic SRVD, fine-tune on CRVD scenes 1–6, and test exclusively on scenes 7–11. We report PSNR, SSIM, and CPU runtime as the main evaluation criteria. The CPU platform is Intel Core i9-10900X at 3.70GHz and the software stack is PyTorch 2.0.1 with OpenCV 4.12.0. To ensure a fair timing comparison, all our methods and baselines are benchmarked using a single CPU thread. FilterFusion timing includes the complete pipeline in NV12 (4:2:0 semi-planar YUV) format: RGB${\to}$NV12 conversion, network inference on the Y plane, filter-bank synthesis on the Y plane and on the $1/4$-resolution UV planes, and NV12${\to}$RGB conversion at the end. The filter-bank kernels (Gaussian, median, weighted sum) are implemented in compiled C with OpenCV's resize for interpolation---a level of optimization that is straightforward for these simple, regular operations but not applicable to dense learned convolutions.

To keep the manuscript rigorous, we only report experiment settings for which complete training and evaluation records are available. Exploratory runs with incomplete logs or provisional metrics are excluded. We compare against two groups of baselines: classical denoisers with predictable runtime behavior, and deep learning baselines trained on the same data.

\subsection{Comparison with State-of-the-Art Methods}

Table~\ref{tab:main} summarizes the full CPU-side comparison. All deep learning baselines follow the same two-stage protocol as the proposed model: pre-training on SRVD synthetic data followed by fine-tuning on real CRVD data. The proposed model reaches 34.72 dB / 0.9330 while requiring dramatically less compute (0.51 total GFLOPs) than all deep learning baselines.
\begin{table}[!t]
    \centering
    \caption{Comparison on CRVD at $1920{\times}1080$, sorted by runtime within each group. All timings are on a \emph{single CPU thread} (Intel i9-10900X) for our methods and baselines. }
    \label{tab:main}
    \scriptsize
    \setlength{\tabcolsep}{2.5pt}
    \renewcommand{\arraystretch}{1.10}
    \begin{tabular}{lccccc}
        \hline
        Method & Params & PSNR $\uparrow$ & SSIM $\uparrow$ & Time (ms) $\downarrow$ & GFLOPs $\downarrow$ \\
        \hline
        \multicolumn{6}{c}{\textit{Real-time methods ($\geq$30\,FPS)}} \\
        \hline
        Gaussian blur & -- & 32.54 & 0.8713 & 7$^\star$ & $\sim$0.06 \\
        Box filter & -- & 31.61 & 0.8822 & 9$^\star$ & $\sim$0.04 \\
        \textbf{FilterFusion} & \textbf{109K} & \textbf{34.72} & \textbf{0.9330} & \textbf{17.9}$^{\star\ddagger}$ & \textbf{0.18}$^\dagger$ \\
        Median filter & -- & 32.48 & 0.8779 & 21$^\star$ & $\sim$0.16 \\
        \hline
        \multicolumn{6}{c}{\textit{Non-real-time methods ($<$30\,FPS)}} \\
        \hline
        TemporalAvg & -- & 17.11 & 0.6815 & 84 & $\sim$0.05 \\
        TemporalAvg + Bilateral & -- & 17.11 & 0.7054 & 125 & $\sim$1.1 \\
        Guided filter~\cite{he2013guided} & -- & 30.30 & 0.8909 & 163 & $\sim$0.22 \\
        Bilateral filter & -- & 33.40 & 0.8926 & 187 & $\sim$1.0 \\
        IterBilateral & -- & 33.26 & 0.8828 & 234 & $\sim$3.0 \\
        EMVD~\cite{maggioni2021emvd} & 41.4K & 38.27 & 0.9722 & 283$^{c}$ & 5.38 \\
        TemporalMedian & -- & 16.66 & 0.6909 & 610 & $\sim$0.12 \\
        AnisoDiffusion~\cite{perona1990scale} & -- & 33.02 & 0.8622 & 1,872 & $\sim$5.7 \\
        SimpleCNN\_5L & 29.5K & 33.78 & 0.9153 & 2,195 & 60.9 \\
        DepthwiseCNN & 5.8K & 32.88 & 0.8750 & 2,234 & 11.7 \\
        MultiFrameCNN$^{a}$ & 34.7K & 31.54 & 0.7697 & 2,400 & 71.7 \\
        NLMeans~\cite{buades2005nonlocal} & -- & 33.22 & 0.9301 & 2,615 & $\sim$134.4 \\
        UNetTiny & 261K & 35.37 & 0.9461 & 4,519 & 122.6 \\
        ReMoNet~\cite{xiang2022remonet} & 809K & 40.12 & 0.9820 & 7,651 & 147.63 \\
        FastDVDnet~\cite{tassano2020fastdvdnet} & 2.48M & 39.91 & 0.9812 & 30,867 & 664.99 \\
        RViDeNet~\cite{yue2020supervised} & 8.56M & 40.03 & 0.9802 & ${\geq}$109,294$^{b}$ & 1,965.04 \\
        Shift-Net~\cite{li2022shiftnet} & 13.38M & 41.00 & 0.9856 & ${\sim}$797K & 2,664.75 \\
        \hline
    \end{tabular}
    \vspace{1mm}
    \begin{flushleft}
    \scriptsize
    $\star$Marks CPU real-time methods ($\geq$30\,FPS).\\
    $\dagger$GFLOPs is network-only (0.18); total including filter-bank synthesis is ${\approx}$0.51\,GFLOPs.\\
    $\ddagger$FilterFusion timing uses compiled C filter-bank synthesis (+ cv2.resize).\\
    $^{a}$CRVD-only (no SRVD phase\,1); SRVD lacks aligned temporal neighbors for multi-frame training.\\
    $^{b}$Lower bound: CUDA-only ops (DCNv2, CrissCrossAttention) replaced by Conv2d.\\
    $^{c}$Single-thread timing, for protocol consistency with all other rows; EMVD's multi-threaded runtime is 128\,ms (discussed in text).\\
    \end{flushleft}
\end{table}

\begin{figure}[!t]
    \centering
    \includegraphics[width=0.48\textwidth]{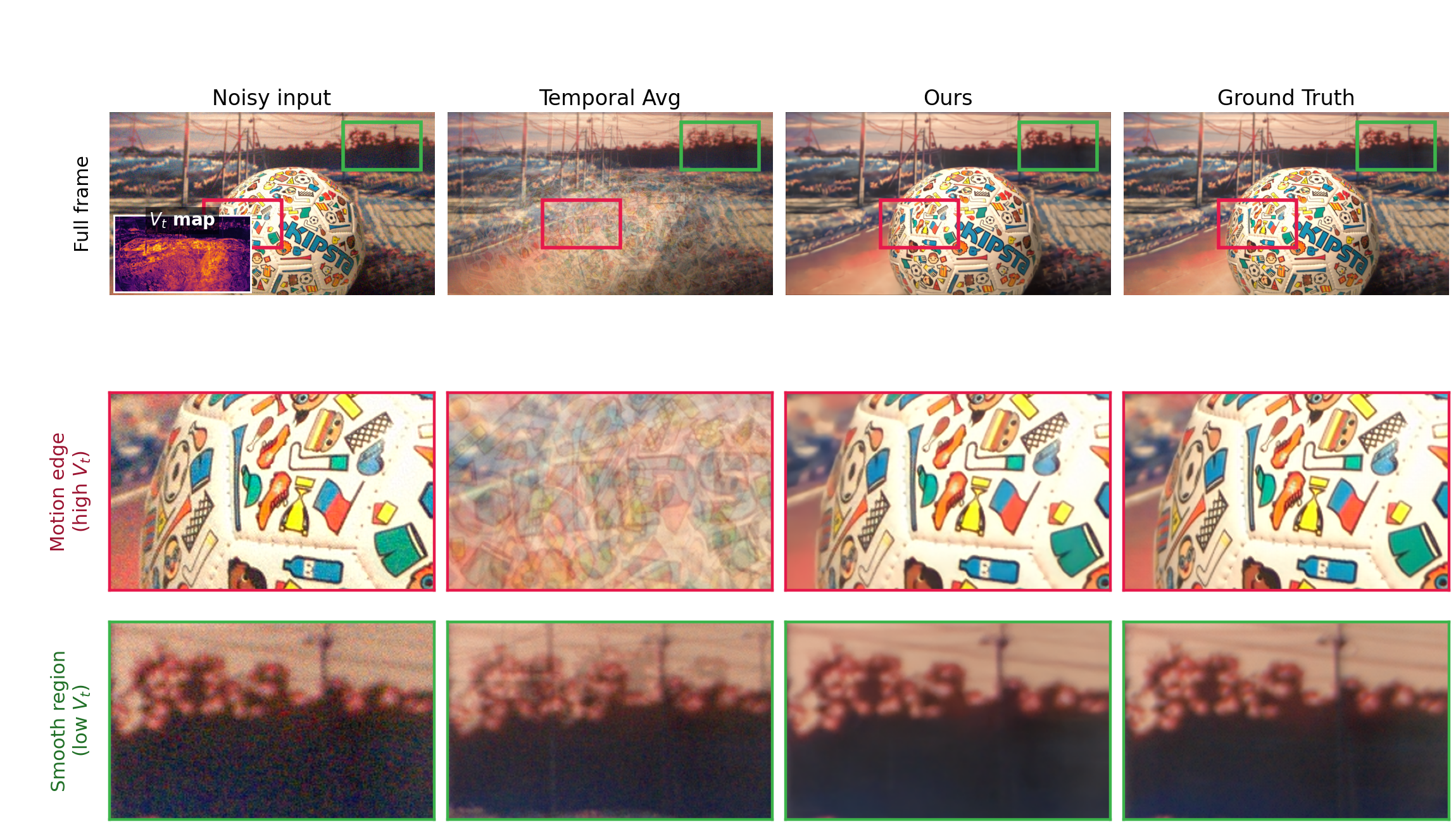}
    \caption{Role of $V_t$ gating on CRVD scene7/ISO12800. Inset (top-left) is the predicted $V_t$ map (warm $=$ high, cool $=$ low). On high-$V_t$ regions (red box), naive multi-frame averaging ghosts the KIPSTA logos while FilterFusion preserves edges. On low-$V_t$ regions (green box), FilterFusion suppresses sensor noise without blurring the thin power lines.}
    \label{fig:qual}
\end{figure}

\textbf{Role of Temporal Variance Gating.}
Figure~\ref{fig:qual} shows how the predicted $V_t$ map gates FilterFusion within one inference pass. On the rotating ball (high $V_t$, red box), naive temporal averaging ghosts the KIPSTA logos---the failure mode that pushes TemporalAvg/Median to ${\sim}$17\,dB in Table~\ref{tab:main}---while FilterFusion reverts to a near single-frame response. On the static sky and power lines (low $V_t$, green box), the same model exploits the multi-frame signal to suppress sensor noise without blurring the wires.

\textbf{Runtime-Quality Analysis.}
Several key observations emerge from Table~\ref{tab:main}.

\textit{Surpassing the classical ceiling.}
Classical single-frame methods plateau around 33.4\,dB regardless of complexity: Bilateral filter achieves 33.40\,dB in 187\,ms, while the far more expensive NLMeans (2,615\,ms, ${\sim}$134\,GFLOPs) and AnisoDiffusion (1,872\,ms) fail to exceed this ceiling. Guided filter scores only 30.30\,dB because its self-guided regime lets the noisy input corrupt the linear coefficient estimation. Naive multi-frame fusion without motion compensation is even worse: TemporalAvg, TemporalMedian, and TempAvg+Bilateral all drop to ${\sim}$17\,dB due to severe ghosting, underscoring the need for learned temporal weighting. FilterFusion breaks through the classical ceiling by +1.32\,dB over the best classical result (Bilateral, 33.40\,dB).

\textit{Speed-quality trade-off vs.\ lightweight CNNs.}
Full-resolution lightweight CNNs are not actually lightweight on CPU at 1080p: despite small parameter counts (5.8K--261K), they require 2.2--4.5\,s per frame on a single thread even after two-stage training (SRVD pre-train + CRVD fine-tune). UNetTiny reaches 35.37\,dB---surpassing FilterFusion's 34.72\,dB---but at the cost of $252{\times}$ slower inference (4,519\,ms vs.\ 17.9\,ms) and $240{\times}$ more compute. This illustrates a key property of our design: FilterFusion trades a modest quality gap ($-$0.65\,dB vs.\ UNetTiny) for a massive speed advantage, making it the only learned method that can operate above real-time at 1080p. With compiled filter-bank synthesis, FilterFusion achieves 17.9\,ms per frame (55.9\,FPS)---$122{\times}$ faster than SimpleCNN\_5L and $252{\times}$ faster than UNetTiny---with only 0.51\,GFLOPs total. We emphasize that this speedup is \emph{architectural}: the filter-bank operations (3${\times}$3 Gaussian, 5${\times}$5 median, weighted sum) are simple enough to admit efficient compiled implementations, whereas CNN baselines already run on highly optimized BLAS/MKL backends and cannot be similarly accelerated. The speed advantage directly validates the core design: low-resolution weight prediction combined with full-resolution filter-bank synthesis avoids the bottleneck of dense learned convolutions at 1080p.

\textit{Extreme efficiency gap to state-of-the-art.}
Published high-accuracy methods achieve 38--41\,dB but at dramatically higher cost. The lightest among them, EMVD (38.27\,dB), requires 5.38\,GFLOPs---$30{\times}$ more than FilterFusion's network cost (0.18\,GFLOPs) and still $10{\times}$ more than our total cost including filter-bank synthesis (0.51\,GFLOPs). This reveals a large, previously unoccupied region in the quality-efficiency landscape: FilterFusion is the only method that simultaneously surpasses the classical quality ceiling while operating at sub-1\,GFLOPs total cost, where no prior learned method exists. 

\textit{Measured CPU runtime of published methods.}
To move beyond GFLOPs-only comparisons, we profiled several published methods on the same CPU platform at $1920{\times}1080$ (Table~\ref{tab:main}, Time column). The results confirm that high-accuracy methods are computationally heavy: FastDVDnet requires ${\sim}$11\,s per frame, RViDeNet at least 43\,s (lower bound, as two CUDA-only operators were replaced by standard convolutions), and Shift-Net exceeds 374\,s. EMVD is a notable case: despite reporting 5.38\,GFLOPs ($30{\times}$ our network cost), it achieves 128\,ms on CPU with multi-threading. FilterFusion's optimized variant reaches 17.9\,ms on a \emph{single} thread---$7.2{\times}$ faster than EMVD's multi-threaded result---highlighting that the filter-bank paradigm is not only compute-efficient but also practically fast because its constituent operations (fixed-kernel convolutions, median, weighted sum) are inherently simple and admit efficient native implementations.

Fig. ~\ref{fig:radar} visualizes these multi-dimensional trade-offs across seven representative methods. The dashed arc marks the real-time boundary (30\,FPS): FilterFusion (55.9\,FPS) and Median filter exceed it, but Median's polygon collapses on quality and structure. FilterFusion achieves the most balanced polygon, with competitive scores on all five axes and the highest quality among real-time-capable methods. Classical methods (Bilateral, NLMeans, Median) score high on lightness (zero parameters) but are limited in quality and---in the case of NLMeans---extremely slow. UNetTiny matches EMVD-level quality (35.37\,dB) and structure (0.9461), but both methods collapse on Speed and Efficiency: UNetTiny requires 4.5\,s per frame and 122.6\,GFLOPs, while EMVD needs 5.38\,GFLOPs and 128\,ms even with multi-threading. This highlights the central trade-off: higher quality is achievable through heavier models, but FilterFusion occupies a unique position as the only method that simultaneously exceeds the real-time threshold and surpasses the classical quality ceiling.

\subsection{Ablation Study}

To validate each design choice, we conduct component-level ablations on CRVD. Table~\ref{tab:ablation} reports PSNR and SSIM for each variant. All ablation models are trained with the same schedule and hyperparameters as the full model unless stated otherwise.

\begin{table}[!t]
    \centering
    \caption{Ablation study on CRVD. Each row disables or modifies one component relative to the full model. All variants are trained \emph{single-stage} on the same data split as the full model in this table (no SRVD pre-training), with the same schedule (160K iterations, Adam, same LR milestones); they are used to attribute quality to specific design choices and are not directly comparable to the two-stage main-table numbers.}
    \label{tab:ablation}
    \footnotesize
    \setlength{\tabcolsep}{4pt}
    \renewcommand{\arraystretch}{1.12}
    \begin{tabular}{clcc}
        \hline
        \# & Variant & PSNR $\uparrow$ & SSIM $\uparrow$ \\
        \hline
        (a) & Full model (FilterFusion) & 34.7549 & 0.9353 \\
        \hline
        \multicolumn{4}{l}{\textit{Guidance signals}} \\
        (b) & w/o temporal variance $V_t$ & 34.8418 & 0.9336 \\
        (c) & w/o spatial variance $V_s$ & 34.6080 & 0.9304 \\
        (d) & w/o neither $V_t$ nor $V_s$ & 34.6354 & 0.9316 \\
        \hline
        \multicolumn{4}{l}{\textit{Filter-bank composition}} \\
        (e) & Raw frames only (no spatial filters) & 32.5283 & 0.8183 \\
        (f) & w/o Gaussian-pyramid branches & 33.5668 & 0.8909 \\
        \hline
        \multicolumn{4}{l}{\textit{Input \& loss}} \\
        (g) & Single frame (no temporal input) & 25.4074 & 0.9043 \\
        (h) & L1 only (no Charb/Sobel/Freq) & 34.8278 & 0.9342 \\
        \hline
    \end{tabular}
\end{table}

\begin{figure}[!t]
    \centering
    \includegraphics[width=0.48\textwidth]{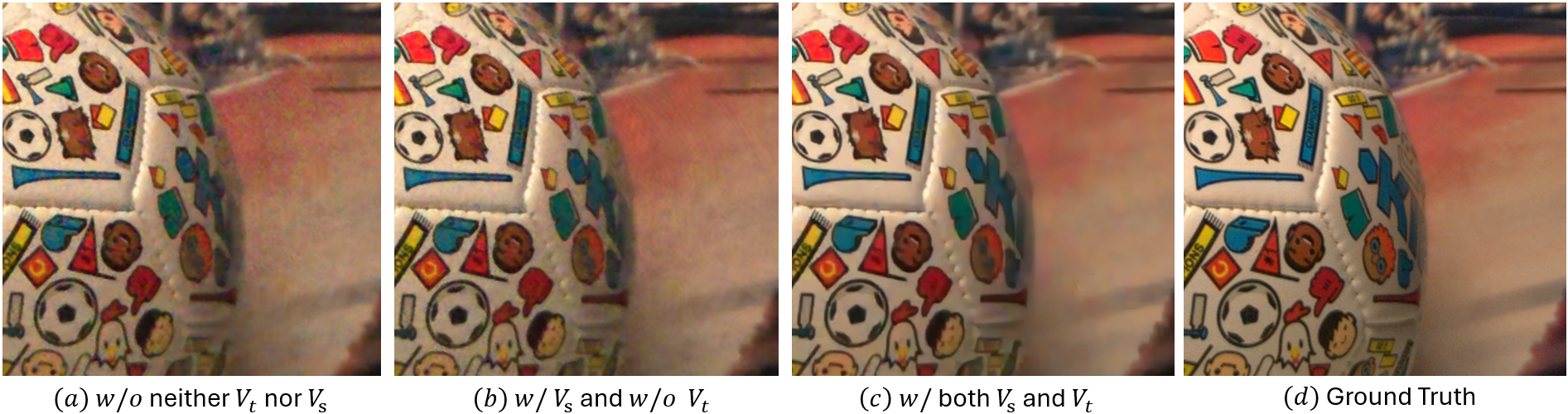}
    \caption{Qualitative ablation on CRVD ISO\,25600 (strong noise, high $V_t$): (a) remains heavy residual noise; (b) noise is cleaned but visible ghosting along the ball's moving edge; (c) clean and ghost-free; (d) ground truth.}
    \label{fig:ablation_vis}
\end{figure}

\textbf{Guidance signals (b--d).}
Removing spatial variance $V_s$ hurts edge-texture preservation, since the low-resolution predictor can no longer distinguish flat-noise regions from structured regions after $8{\times}$ downsampling (Fig.~\ref{fig:ablation_vis}a). Removing temporal variance $V_t$ degrades quality in motion regions, where the controller loses its explicit cue about temporal reliability and causes ghost artifacts (Fig.~\ref{fig:ablation_vis}b). Quantitatively, $V_s$ contributes more than $V_t$ on PSNR ($-0.15$ vs.\ $+0.07$\,dB), but $V_t$ is decisive for SSIM ($-0.0017$): without $V_t$, the controller over-trusts multi-frame averaging---residual static-background noise drops (raising PSNR), but motion ghosts at the ball's edge violate local structure, a distortion SSIM's local-window metric catches while averaged PSNR hides.

\textbf{Filter-bank composition (e, f).}
Variant~(e) reduces the filter bank to only the raw input frames ($K{=}2$: temporal mean and reference), removing all spatial filters (Gaussian pyramid and median). This tests whether the multi-filter synthesis---the central contribution of our design---adds value beyond simple temporal fusion. Variant~(f) selectively removes the Gaussian-pyramid branches while keeping the median filter, testing the importance of multi-scale smoothing. Removing all spatial filters~(e) drops PSNR by 2.23\,dB and SSIM by 0.117---the largest drop among ablations after~(g); removing only the pyramid~(f) costs 1.19\,dB and 0.044, confirming multi-scale smoothing contributes a substantial portion of the gain.

\textbf{Input and loss (g, h).}
The single-frame variant~(g) collapses to 25.41\,/\,0.9043 ($-9.35$\,dB vs.\ full), confirming that temporal fusion is the dominant source of quality.  Variant~(h) replaces the combined loss $\mathcal{L}$ with single $\mathcal{L}_{\mathrm{L1}}$ and yields slightly higher PSNR but lower SSIM (34.83\,/\,0.9342). The result demonstrates that $\mathcal{L}_{\mathrm{charb}} + \mathcal{L}_{\mathrm{sobel}} + \mathcal{L}_{\mathrm{freq}}$ jointly provide complementary structure and spectrum aware constraints, suppressing residual high-frequency noise and improving structural fidelity beyond pixel-domain supervision.

\subsection{Discussion}

FilterFusion is designed for practical deployment as much as for accuracy. Because the branch semantics are explicit, the system can be simplified by dropping selected branches or by replacing the temporal backbone with a lighter alternative while keeping the overall fusion formulation unchanged. Without explicit motion estimation, temporal-variance guidance steers unreliable regions toward reference-preserving branches to suppress ghosting, so highly dynamic regions receive less temporal denoising as a deliberate quality–latency trade-off.

The filter-bank operations are \emph{simple and regular}---fixed-kernel spatial filters and element-wise weighted sums---which map naturally to efficient compiled code. In contrast, CNN baselines already execute on highly optimized BLAS/MKL backends within PyTorch; their dense learned convolutions leave little room for further implementation-level gains. The pipeline breakdown at 1080p is: network inference 2.2\,ms, Y-channel post-processing 11.5\,ms, UV post-processing 4.2\,ms, totaling 17.9\,ms (55.9\,FPS)---demonstrates that the 0.51\,GFLOPs total budget translates directly into real-time throughput on a \emph{single} CPU thread.

\section{Conclusion}
\label{sec:conclusion}

We presented FilterFusion, a prior-guided adaptive fusion framework for efficient video denoising. The model predicts a compact low-resolution guidance signal and synthesizes the full-resolution output through a small set of complementary denoising responses. Experiments on CRVD (sensor noise) show that this structured design surpasses the classical quality ceiling (34.72\,dB vs.\ 33.40\,dB for Bilateral) while using orders of magnitude less compute than CNN alternatives ($240{\times}$ less than UNetTiny). These results support the central claim of the paper: for deployment-oriented video denoising, low-resolution control combined with explicit high-resolution filter synthesis is a practical and effective design principle. With compiled native filter-bank synthesis, FilterFusion achieves 17.9\,ms per frame (55.9\,FPS) at 1080p on a single CPU thread---comfortably exceeding real-time without multi-threading, and $7.2{\times}$ faster than EMVD's multi-threaded result. This is enabled by the architectural choice: the filter-bank operations are simple enough to admit efficient compiled implementations, a property not shared by dense learned convolutions.

\bibliographystyle{IEEEtran}
\bibliography{references}

\end{document}